\documentclass{article} 
\pdfoutput=1
\usepackage{iclr_style_mod,times}

\iclrfinalcopy
\usepackage{aas_macros}
\newcommand{\anonfinal}[2]{#2} 

\title{\vspace{-26pt}Transformation Importance with Applications to Cosmology}
\newcommand{\hneg}{\hspace{-18pt}}

\author{\hneg Chandan Singh\textsuperscript{1,*}, Wooseok Ha\textsuperscript{2,*}, Fran\c cois Lanusse\textsuperscript{3,4}, Vanessa Boehm\textsuperscript{3}, Jia Liu\textsuperscript{3}, Bin Yu\textsuperscript{1, 2, 5, 6, 7} \vspace{3pt} \\ 
    \hneg \textsuperscript{1}EECS Department, UC Berkeley \hspace{21pt}
    \hneg \textsuperscript{2}Statistics Department, UC Berkeley\\
    \hneg \textsuperscript{3}Berkeley Center for Cosmological Physics, UC Berkeley and LBNL\\
    \hneg \textsuperscript{4}Laboratoire AIM, CEA, CNRS,\\
    Universit\`e Paris-Saclay, Universit\`e Paris Diderot, Sorbonne Paris Cit\`e\\
    \hneg \textsuperscript{5}Chan Zuckerberg Biohub \hspace{10pt} \textsuperscript{6}Center for Computational Biology, UC Berkeley\\
    \hneg \textsuperscript{7}Environmental Genomics and Systems Biology Division, LBNL\\
    \hneg \textsuperscript{*}Denotes equal contribution \vspace{3pt}\\
    \hneg \texttt{\{cs1,haywse,flanusse,vboehm,jialiu,binyu\}@berkeley.edu}
    \vspace{-8pt}}

\date{} 
\usepackage{graphicx}
\usepackage{float}

\usepackage{titling}
\usepackage{amsmath}
\usepackage{amssymb}
\usepackage{changepage}

\usepackage{placeins} 
    
\newcommand{\fref}[1]{Fig~\ref{#1}}

\newcommand{\sref}[1]{Sec~\ref{#1}}
\newcommand{\tref}[1]{Table~\ref{#1}}

\newtheorem{definition}{Definition} 

\definecolor{cadetblue}{rgb}{0.37, 0.62, 0.63}
\definecolor{cerulean}{rgb}{0.0, 0.48, 0.65}
\definecolor{darkcerulean}{rgb}{0.03, 0.27, 0.49}
\usepackage[citecolor=cerulean, linkcolor=darkgray, urlcolor=darkcerulean, colorlinks=true]{hyperref}
\usepackage[noadjust]{cite}
\newcommand{\citep}[1]{\cite{#1}}
\begin{document}

\maketitle

\begin{abstract}
Machine learning lies at the heart of new possibilities for scientific discovery, knowledge generation, and artificial intelligence. Its potential benefits to these fields requires going beyond predictive accuracy and focusing on interpretability.
In particular, many scientific problems require interpretations in a domain-specific interpretable feature space (e.g. the frequency domain) whereas attributions to the raw features  (e.g. the pixel space) may be unintelligible or even misleading.
To address this challenge, we propose TRIM (\underline{Tr}ansformation \underline{Im}portance), a novel approach which attributes importances to features in a transformed space and can be applied post-hoc to a fully trained model.\footnote{\anonfinal{\url{https://github.com/anonymized}}{\url{https://github.com/csinva/transformation-importance}} contains notebooks, scripts, and pre-trained models for reproducing the results here and applying the methods here to new models.} 
TRIM is motivated by a cosmological parameter estimation problem using deep neural networks (DNNs) on simulated data, but it is generally applicable across domains/models and can be combined with any local interpretation method.
In our cosmology example, combining TRIM with contextual decomposition \citep{murdoch2018beyond} shows promising results for identifying which frequencies a DNN uses, helping cosmologists to understand and validate that the model learns appropriate physical features rather than simulation artifacts.
\end{abstract}


\section{Introduction and Related Work}
\label{sec:intro}

Due to its impressive predictive performance, machine learning has established itself as a crucial tool across a variety of domains.
In scientific fields, where interpretation is critical, interpreting these models is a key next step for scientific discovery.
As a result, the field of interpretable machine learning has become increasingly important \cite{murdoch2019definitions, doshi2017roadmap, molnar2019interpretable}. 
Thus far, a large majority of interpretability work has focused on attributing importance to raw features, such as pixels in an image or words in a document \citep{sundararajan2016gradients, selvaraju2016grad, ribeiro2016should, shrikumar2016not, dabkowski2017real, devlin2019disentangled, petsiuk2018rise, baehrens2010explain, bach2015pixel, zintgraf2017visualizing}, 
with many similarities among the methods \citep{ancona2018towards, lundberg2017unified}. However, when features are highly correlated or features in isolation are not semantically meaningful, the resulting attributions need to be improved.

To meet this challenge, we propose TRIM (\underline{Tr}ansformation \underline{Im}portance), an approach for attributing importance to transformations of the input features (see \fref{fig:t}). This is critical for making interpretations relevant to a particular audience/problem, as attributions in a domain-specific feature space (e.g. frequencies or principal components) can often be far more interpretable than attributions in the raw feature space (e.g. pixels or biological readings). Moreover, features after transformation can be more independent, semantically meaningful, and comparable across data points. This idea is related to existing works suggesting the use of a ``simplified input-representation'' \citep{ribeiro2016should, lundberg2017unified}, but we generalize these works beyond transformations which map existing features into simplified binary features.
The work here focuses on combining TRIM with contextual decomposition (CD), an existing attribution method \citep{murdoch2018beyond}, although TRIM can be combined with any local interpretation method.


We focus on  cosmology example, where attributing importance to transformations helps understand cosmological models in a more interpretable feature space. 
Specifically, we consider weak gravitational lensing convergence maps, i.e. maps of the mass distribution in the Universe integrated up to a certain distance from the observer. In a cosmological experiment (e.g. a galaxy survey), these mass maps are obtained by measuring the distortion of distant galaxies caused by the deflection of light by the mass between the galaxy and the observer~\citep{Bartelmann2001}. These maps contain a wealth of physical information of interest to cosmologists, such as the total matter density in the universe, $\Omega_m$.
Current research aims at identifying the most informative features in these maps for inferring the true cosmological parameters. 
The traditional summary statistic for lensing maps is the power spectrum which is known to be sub-optimal for parameter inference. Tighter parameter constraints can be obtained by including higher-order statistics, such as the bispectrum~\citep{Coulton2019} and peak counts~\citep{LiuPeaks2019}. However, DNN-based inference methods claim to improve on constraints based on these traditional summaries~\citep{Zorilla2019, ribli2019improved, Fluri2019}.

On top of the accurate predictive power of a DNN, here it is also important to understand what the model learns. Knowing which features are important provides deeper understanding and can be used to design optimal experiments or analysis methods. Moreover, because these models are trained on numerical simulations (realizations of the Universe with different cosmological parameters), it is important to validate that the model uses physical features rather than latching on to numerical artifacts in the simulations. TRIM shows promise for understanding and validating that the DNN learns appropriate physical features by analyzing attributing importance in the spectral domain.




\section{Calculating transformation importance}

\begin{figure}[hbtp]
    \centering
    \includegraphics[width=0.4\textwidth]{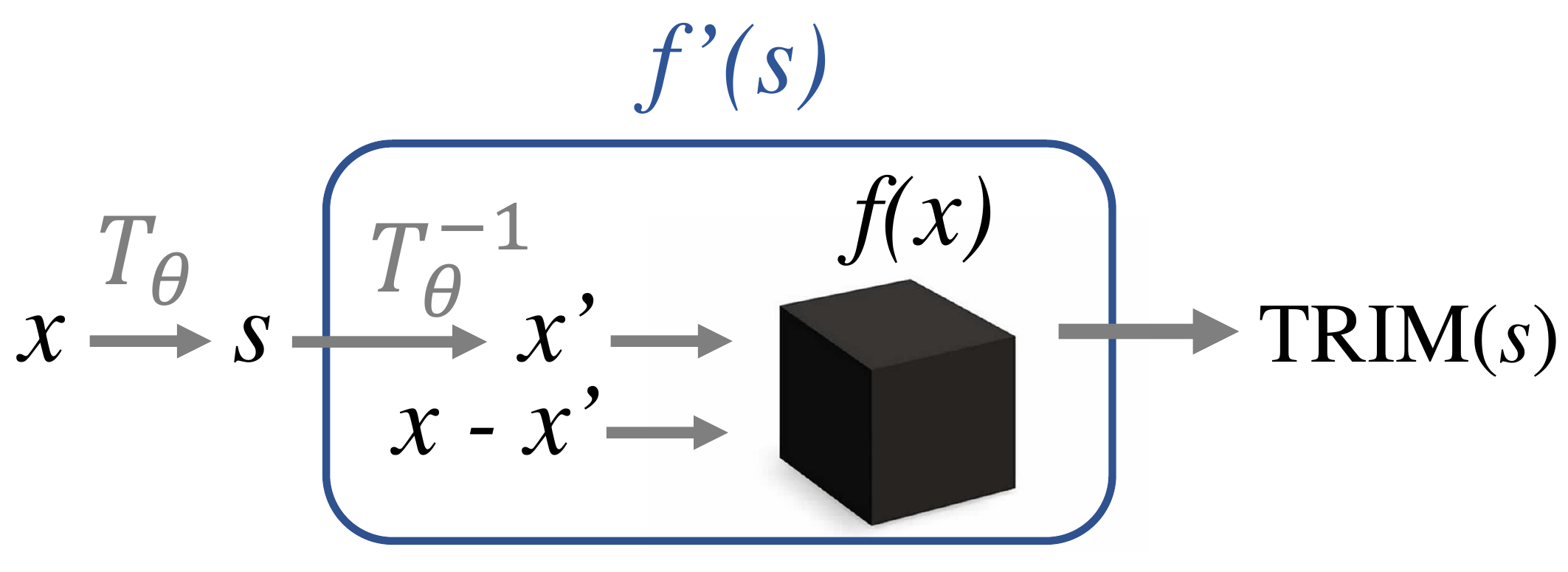}
    \caption{TRIM: Attributing importance to a transformation of an input $T_\theta(x)$ given a model $f(x)$.}
    \label{fig:t}
\end{figure}

We aim to interpret the prediction made by a model $f$ given a single input $x$.
The input $x$ is in some domain $\mathcal X$, but we desire an explanation for its representation $s$ in a different domain $\mathcal S$, defined by a mapping $T: \mathcal X \to \mathcal S$, such that $s = T (x)$. 
For example, if $x$ is an image, $s$ may be its Fourier representation, and $T$ would be the Fourier transform. Notably, this process is entirely post-hoc: the model $f$ is already fully trained on the domain $\mathcal X$. By reparametrizing our network as shown in \fref{fig:t}, we can obtain attributions in the domain $\mathcal S$. If we require that the mapping $T$ be invertible, so that $x = T^{-1} (s)$, we can represent each data point $x$ with its counterpart $s$ in the desired domain, and our function to interpret becomes $f' = f \circ T^{-1}$; the function $f'$ can be interpreted with any existing local interpretation method $attr$ (e.g. LIME \citep{ribeiro2016should}, Integrated Gradients \citep{sundararajan2016gradients})).\footnote{Note that if the transformation $T$ is not perfectly invertible (i.e. $x \neq x'$), then the residuals $x-x'$ may also be required for local interpretation. For example, they are required for any gradient-based attribution method to aid in computing $\partial f' / \partial s$.}
Once we have the reparameterized function $f'(s)$, we need only specify which part of the input to interpret to define TRIM:

\begin{definition} Given a model $f$, an input $x$, a mask $M$, a transformation $T$, and an attribution method $attr$, 
\label{def_t}
\vspace{2pt}
\begin{align*}
    \textup{TRIM}(s) = attr\left(f'; s \right) \\
    \textup{where } f' = f \circ T^{-1},\; s = M \odot T(x) 
\end{align*}
\vspace{2pt}
Here $M$ is a mask used to specify which parts of the transformed space to interpret and $\odot$ denotes elementwise multiplication.
\end{definition}

In the work here, the choice of attribution method $attr$ is CD, as it can disentangle the importance of features and their interactions, and has been rigorously evaluated using real data \citep{murdoch2018beyond}, human experiments \citep{singh2018hierarchical}, and during model training \citep{rieger2019interpretations}. 
In this case, $attr\left(f; x', x \right)$ represents the CD score for the features $x'$ as part of the input $x$.
Different from previous work, this formulation does not require that $x'$ simply be a binary masked version of $x$. Rather, the selection of the mask $M$ allows a human/domain scientist to decide which transformed features to score. In the case of image classification, rather than simply scoring a pixel, one may score the contribution of a frequency band to the prediction $f(x)$. In this case, $T$ is the FFT and $M$ is a mask which is zero for frequencies outside of the band and one for frequencies inside of the band, so that $x'$ represents the bandpass-filtered image.

This general setup allows for attributing importance to a wide array of transformations. For example, $T$ could be any invertible transform (e.g. a wavelet transform), or a linear projection (e.g. onto a sparse dictionary). Moreover, we can parameterize the transformation $T_\theta$ and learn the parameters $\theta$ to produce a desirable representation (e.g. sparse or disentangled). 
\section{Qualitative examples}

\begin{figure}[hbtp]
    \centering
        \includegraphics[width=0.8\textwidth]{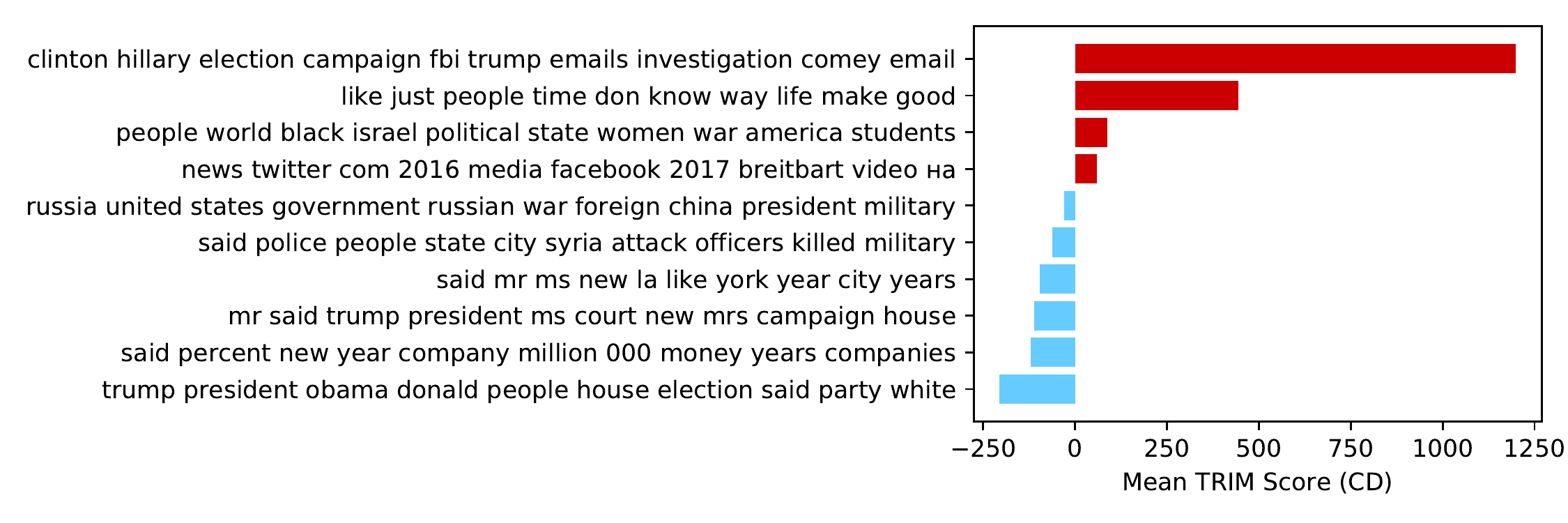}
        
    \caption{TRIM attributions for a fake-news classifier based on a topic model transformation. Each row shows one topic, labeled with the top ten words in that topic. Higher attributions correspond to higher contribution to the class \textit{fake}. Calculated over all points which were accurately classified as \textit{fake} in the test set (4,160 points).}
    \label{fig:fakenews_example}
    
\end{figure}

We investigate a text-classification setting using TRIM. We train a 3-layer fully connected DNN with ReLU activations on the Kaggle Fake News dataset\footnote{\url{https://www.kaggle.com/c/fake-news/overview}}, achieving a final test accuracy of 94.8\%. The model is trained directly on a bag-of words representation, but TRIM can provide a more succinct space via a topic model transformation (learned via latent dirichlet allocation \citep{blei2003latent}). \fref{fig:fakenews_example} shows the mean attributions for different topics when the model predicts \textit{Fake}. Interestingly, the topic with the highest mean attribution contain recognizable words such as \textit{clinton} and \textit{emails}.

\section{Cosmology experiments}
\begin{figure}[hbtp]
    \begin{adjustwidth}{-1.2cm}{-1.2cm}
    \centering
    \includegraphics[height=1in]{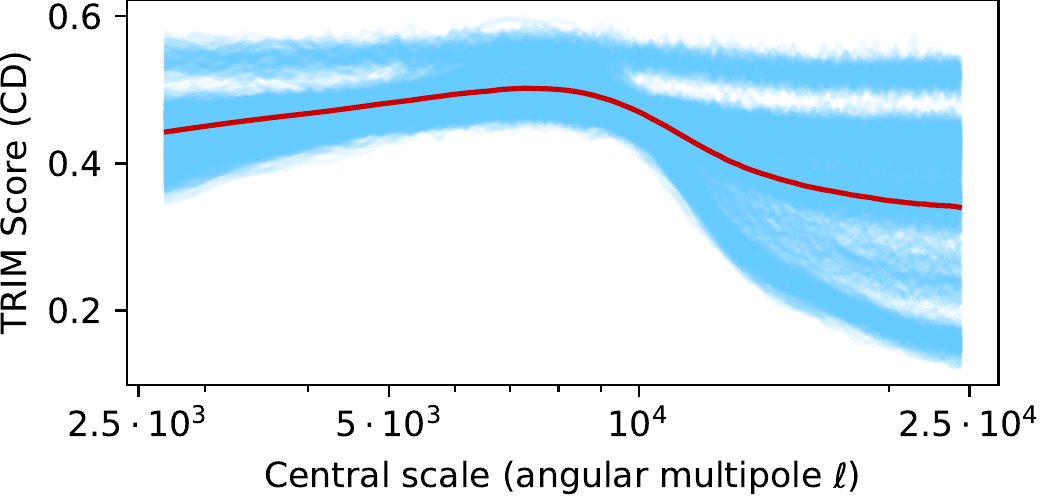}
    \includegraphics[height=1.22in]{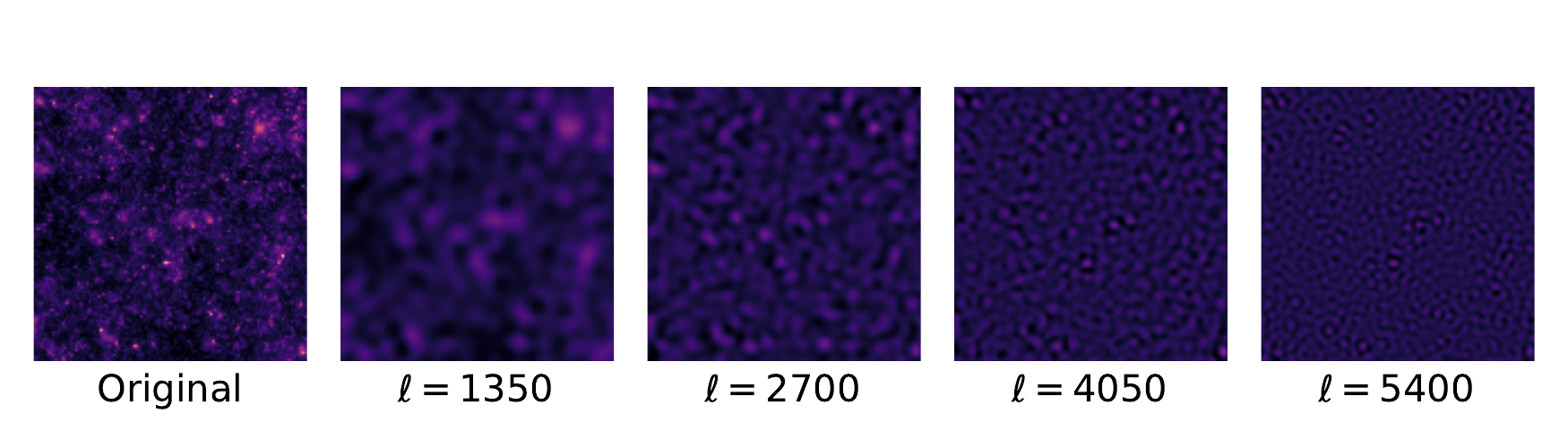}
    \caption{Different scales (i.e. frequency bands) contribute differently to the prediction of $\Omega_m$. Each blue line corresponds to one testing image and the red line shows the mean. Images show the features present at different scales. The bandwidth is $\Delta_\ell=$2,700.}
    \label{fig:fft}
    \end{adjustwidth}
\end{figure}

We now return to the question of interpreting a model trained to predict $\Omega_m$ from simulated weak gravitational lensing convergence maps.
We train a DNN\footnote{The model's architecture is Resnet 18 \citep{he2016deep}, modified to take only one input channel.} to predict $\Omega_m$ from 100,000 mass maps simulated at 10 different cosmologies from the \texttt{MassiveNuS} simulations \citep{Jia2018}, achieving an $R^2$ value of 0.92 on the test set (10,000 mass maps). Full simulation details are given in \sref{sec:simulation details}. To understand what features the model is using, we desire an interpretation in the space of the power spectrum. The images in \fref{fig:fft} show how different information is contained within different frequency bands in the mass maps. The plot in \fref{fig:fft} shows the TRIM attributions (normalized by the predicted value) for different frequency bands when predicting the parameter $\Omega_m$. Interestingly, the most important frequency band for the predictions seems to peak at scales around $\ell=10^4$ and then decay for higher frequencies. Our physical interpretation of this result is that the DNN  concentrates on the most discriminative part of the Power Spectrum, i.e. at scales large enough not to be dominated by sample variance, and smaller than the frequency cutoff at which the simulations lose power due to resolution effects.

\begin{figure}[hbtp]
    \centering
    \includegraphics[width=\textwidth]{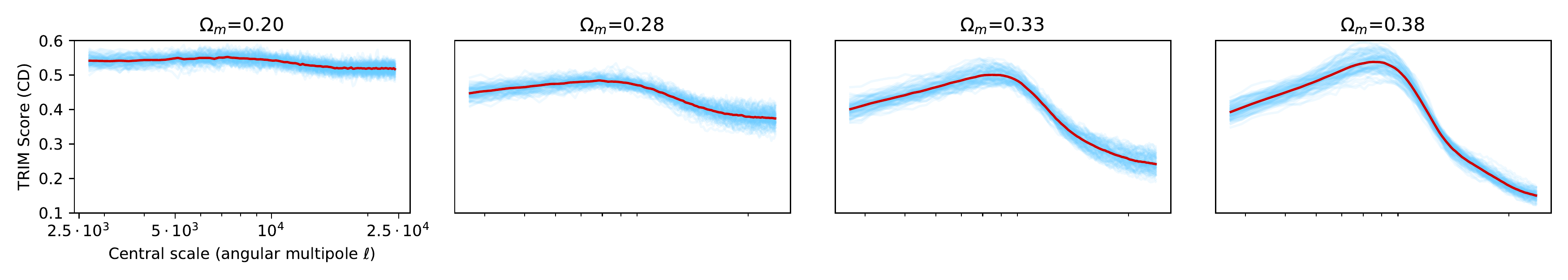}
    \caption{TRIM attributions vary with the value of $\Omega_m$.}
    \label{fig:vary_omegam}
\end{figure}

\fref{fig:vary_omegam} shows some of the curves from \fref{fig:fft} separated based on their cosmology, to show how the curves vary with the value of $\Omega_m$. Increasing the value of $\Omega_m$ increases the contribution of scales close to $\ell=10^4$, making other frequencies relatively unimportant. This seems to correspond to known cosmological knowledge, as these scales seem to correspond to galaxy clusters in the mass maps, which are structures very sensitive to the value of $\Omega_m$.
The fact that the importance of these features vary with $\Omega_m$ would seem to indicate that at lower $\Omega_m$ the model is using a different source of information, not located at any single scale, for making its prediction.

\FloatBarrier
\paragraph{Evaluation via simulation}
In the case of a perfectly invertible transformation, such as the Fourier transform, TRIM simply measures the ability for the underlying attribution method (in this case CD) to correctly attribute importance in the transformed space. As such, we can rely on the careful evaluation of contextual decomposition in previous work, where it has been shown to (1) accurately recover known feature importances and feature interactions \citep{murdoch2018beyond}, (2) correctly inform human decision-making and be robust to adversarial perturbations \citep{singh2018hierarchical}, and (3) reliably alter a neural network's predictions when regularized appropriately \citep{rieger2019interpretations}.

On top of these evaluations, we add synthetic simulations showing the ability of CD to recover known groundtruth feature importances. Features are generated i.i.d. from a standard normal distribution. Then, a binary classification outcome is defined by selecting a random frequency and testing whether that frequency is greater than its median value. Finally, we train a 3-layer fully connected DNN with ReLU activations to learn this classification task and then test the ability of different methods to assign this frequency the highest importance. \tref{tab:simulations} shows the percentage of errors made by different methods in such a setup. CD has the lowest error on average, compared to popular baselines.


\begin{table}[H]

\centering
\small

\begin{tabular}{cccc}
    \hline
    CD  & DeepLift \citep{shrikumar2016not} & SHAP \citep{lundberg2017unified} & Integrated Gradients \citep{sundararajan2016gradients}\\
    \hline
    \textbf{0.4 $\pm$ 0.282}  & 3.6 $\pm$ 0.833     & 4.0 $\pm$ 0.897  & 4.2 $\pm$ 0.876    \\
    \hline
\end{tabular}
    \caption{Error (\%) in recovering a groundtruth important frequency in simulated data using different attribution methods with TRIM, averaged over 500 simulated datasets.}
\label{tab:simulations}
\end{table}
\FloatBarrier

\paragraph{Discussion}
The results here show promise for TRIM to enable deeper understanding in cosmology and suggest potential uses for TRIM across a variety of different domains.
Moreover, the TRIM experiments here can be extended to a much broader class of transformations, which could be selected by a domain expert or optimized to exhibit different desirable properties.
Ultimately, we hope TRIM can contribute to a new wave of scientific discovery using machine learning. 

\section{Acknowledgements}

The authors would like to thank Jamie Murdoch, Alan Dong, and the Yu group for helpful feedback and discussions. This research was supported in part by grants ARO W911NF1710005, ONR N00014-16-1-2664, NSF DMS-1613002, NSF IIS 174134, NSF 82555, NSF 2015341, the NSF TRIPOD program, and by both NSF through Award 2031883 and Simons Foundation through Award 814639 for the Collaboration on the Theoretical Foundations of Deep Learning. We thank the Center for Science of Information (CSoI), a US NSF Science and Technology Center, under grant agreement CCF-0939370, and the Simons Foundation through Award 814639 for the Collaboration on the Theoretical Foundations of Deep Learning.

\FloatBarrier
{
    \footnotesize
    \bibliographystyle{unsrt}

}
\FloatBarrier
\pdfoutput=1

\setcounter{table}{0}
\setcounter{figure}{0}
\setcounter{section}{0}
\renewcommand{\thetable}{S\arabic{table}}
\renewcommand{\thefigure}{S\arabic{figure}}
\renewcommand{\thesection}{S\arabic{section}} 



\begin{center}
    \Huge
    Supplement
\end{center}
\label{sec:supp}

\section{Simulation details}
\label{sec:simulation details}

In this work, we use the publicly available \texttt{MassiveNuS} simulation suite \citep{Jia2018}, composed of 101 different N-body simulations spanning a range of cosmologies varying three parameters: the total neutrino mass $\Sigma m_\nu$, the normalization of the primordial power spectrum $A_s$, and the total matter density $\Omega_m$. These simulations are run at a single resolution of 1024$^3$ particles for a 512 Mpc/$h$ box size, and then ray-traced to obtain lensing convergence maps at source redshifts ranging from $z_s$ =1.0 to $z_s=1100$.
To build our dataset, we select 10 different cosmologies, listed in \autoref{tab:sim_params}, each of which provides 10,000 mass maps at source redshift $z_s=1$. We rebin these maps to size 256x256 with a pixel resolution of 0.8 arcmin. 

\begin{table}[H]
    \centering
    \small
    \begin{tabular}{c|c|c}
        $m_\nu$ & $\Omega_m$ & $10^9A_s$\\
        \hline
         0.0 & 0.3 & 2.1 \\
0.06271 & 0.3815 & 2.2004 \\
0.06522 & 0.2821 & 1.8826 \\
0.06773 & 0.4159 & 1.6231 \\
0.07024 & 0.2023 & 2.3075 \\
0.07275 & 0.3283 & 2.2883 \\
0.07526 & 0.3355 & 1.5659 \\
0.07778 & 0.2597 & 2.4333 \\
0.0803 & 0.2783 & 2.3824 \\
0.08282 & 0.2758 & 1.8292 \\
    \end{tabular}
    \caption{Parameter values used in cosmology simulations.}
    \label{tab:sim_params}
\end{table}

\section{TRIM attribution curves across cosmological parameters}

\begin{figure}[H]
    \centering
    \includegraphics[width=\textwidth]{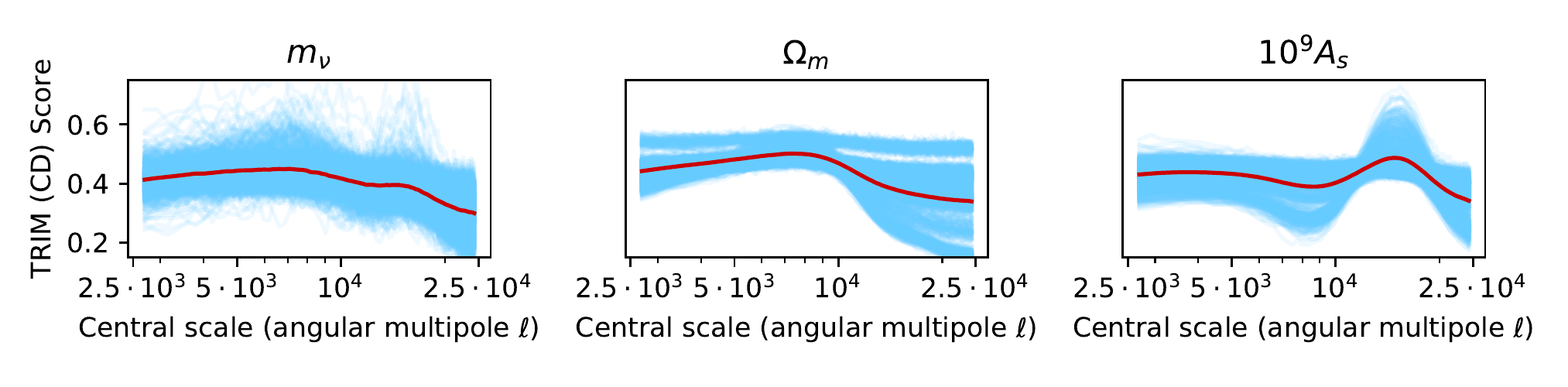}
    \caption{Attribution curves for DNNs predicting cosmological parameters.}
    \label{fig:all_curves}
\end{figure}

\section{More qualitative examples}

\begin{figure}[H]
    \centering
    \includegraphics[width=0.6\textwidth]{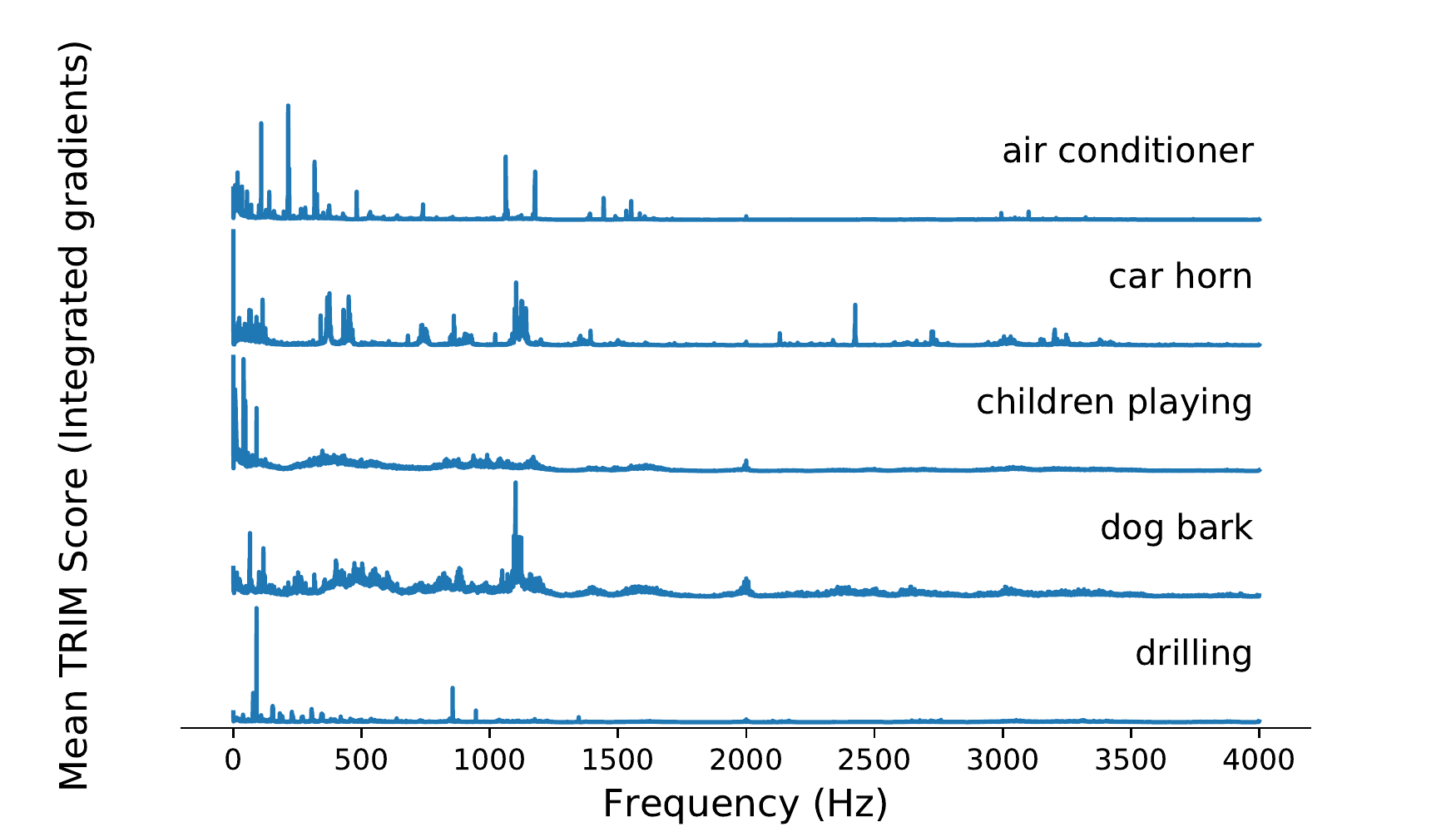}
    \caption{TRIM attributions in the Fourier domain for a DNN trained to perform audio-classification. Each curve shows the importance (IG score) of each frequency to the prediction of the correct class. The model's architecture is based on the M5 architecture \citep{dai2017very}, and it is trained directly on the raw audio waveforms. It achieves 57.2\% top-1 test accuracy. This figure shows the first 5 classes out of 10.}
    \label{fig:audio_example}
\end{figure}

\fref{fig:audio_example} shows an example of attributions for a classifier trained on the UrbanSounds8K audio-classification dataset \citep{salamon2014dataset}.

\begin{figure}[H]
    \centering
    \includegraphics[width=0.6\textwidth]{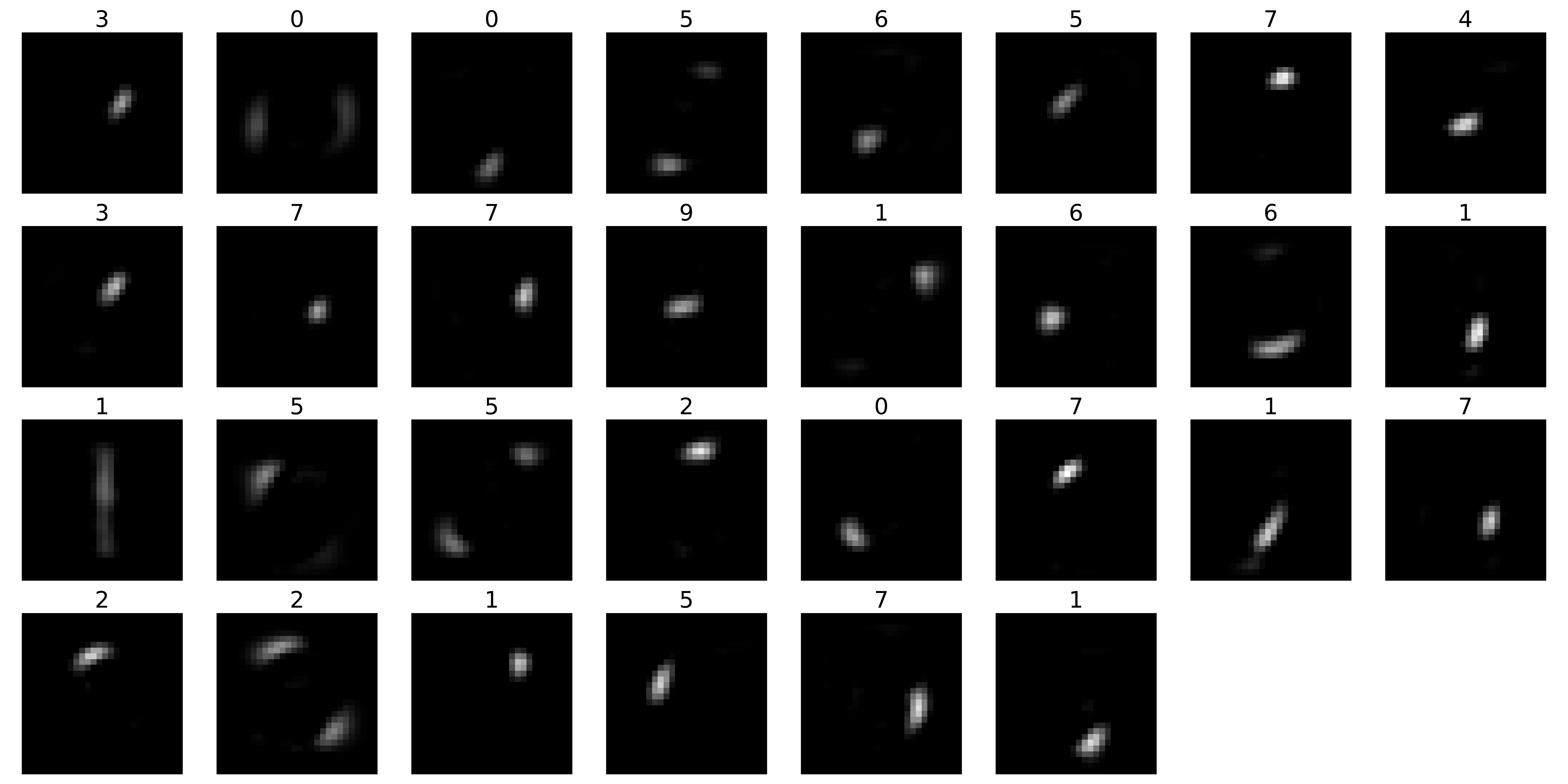}
    \caption{TRIM attributions for an NMF basis for a DNN trained to perform digit-classification. The number above each image shows the class which the interpretation says is most contributed to, based on the mean TRIM score (CD).}
    \label{fig:nmf}
\end{figure}

\fref{fig:nmf} shows an example of attributions for a classifier trained on the MNIST dataset \cite{lecun1998mnist} to perform digit-classification. The DNN is based on a standard architecture\footnote{Retrieved from \url{https://github.com/pytorch/examples/tree/master/mnist}.} and achieves 97.7\% test accuracy.

\section{Parameterizing and learning the transformations}

We can go further with the methods proposed here to learn the transformations to induce some desired properties. More specifically, we can parameterize the transformation $T$ as $T_\theta$, and learn $\theta$ via optimization.


For example, in standard dictionary learning, we may want to learn a dictionary such that the activations are sparse.

\begin{align}
    \theta &= \underset{\theta}{\text{argmin}} ||T_\theta(x)||_1\\
    \text{subject to} &\quad x = T_\theta^{-1} \circ T_\theta(x)
\end{align}

We also know the attributions $\text{TRIM}(s)$. We might want to learn $\theta$ to make the attributions sparse, or spatially contiguous.

\begin{align}
    \theta &= \underset{\theta}{\text{argmin}} ||T_\theta(x)||_1 +  ||\text{TRIM} (T_\theta(x))||_1\\
    \text{subject to} &\quad x = T_\theta^{-1} \circ T_\theta (x)
\end{align}

In this case, the TRIM scores correspond to a fully trained model $f$.








\end{document}